\documentclass[conference]{IEEEtran}
\IEEEoverridecommandlockouts
\usepackage{cite}
\usepackage{amsmath,amssymb,amsfonts}
\usepackage{algorithmic}
\usepackage{graphicx}
\usepackage{textcomp}
\usepackage{xcolor}
\usepackage{hyperref}
\usepackage{gensymb}
\def\BibTeX{{\rm B\kern-.05em{\sc i\kern-.025em b}\kern-.08em
    T\kern-.1667em\lower.7ex\hbox{E}\kern-.125emX}}
\begin{document}

\title{ Polyp detection in colonoscopy images using YOLOv11\\
{\footnotesize \textsuperscript{}}
\thanks{}
}

\author{\IEEEauthorblockN{1\textsuperscript{st} Alok Ranjan Sahoo}
\IEEEauthorblockA{\textit{Department of CSIT, ITER} \\
\textit{SOA University}\\
Bhubaneswar, India\\
aloksahoo@soa.ac.in}
\and
\IEEEauthorblockN{2\textsuperscript{nd} Satya Sangram Sahoo}
\IEEEauthorblockA{\textit{Department of CSIT} \\
\textit{SOA University}\\
Bhubaneswar, India\\
satyasangramsahoo@soa.ac.in}
\and
\IEEEauthorblockN{3\textsuperscript{rd} Pavan Chakraborty}
\IEEEauthorblockA{\textit{Department of IT} \\
\textit{IIIT Allahabad}\\
Prayagraj, India \\
pavan@iiita.ac.in}



}

\maketitle

\begin{abstract}

Colorectal cancer (CRC) is one of the most commonly diagnosed cancers all over the world. It starts as a polyp in the inner lining of the colon. To prevent CRC, early polyp detection is required. Colonosopy is used for the inspection of the colon. Generally, the images taken by the camera placed at the tip of the endoscope are analyzed by the experts manually. Various traditional machine learning models have been used with the rise of machine learning. Recently, deep learning models have shown more effectiveness in polyp detection due to their superiority in generalizing and learning small features. These deep learning models for object detection can be segregated into two different types: single-stage and two-stage. Generally, two stage models have higher accuracy than single stage ones but the single stage models have low inference time. Hence, single stage models are easy to use for quick object detection. YOLO is one of the single-stage models used successfully for polyp detection. It has drawn the attention of researchers because of its lower inference time. The researchers have used Different versions of YOLO so far, and with each newer version, the accuracy of the model is increasing. This paper aims to see the effectiveness of the recently released YOLOv11 to detect polyp. We analyzed the performance for all five models of YOLOv11 (YOLO11n, YOLO11s, YOLO11m, YOLO11l, YOLO11x) with Kvasir dataset for the training and testing. Two different versions of the dataset were used. The first consisted of the original dataset, and the other was created using augmentation techniques. The performance of all the models with these two versions of the dataset have been analysed.

      
\end{abstract}


\section{Introduction}
As per the global cancer statistics 2022\cite{bray2024global}, Colorectal cancer (CRC) is the second most leading cause of cancer death after lung cancer. It is the third most commonly diagnosed cancer after lung cancer and female breast cancer. It accounts for 1,926,118 new cases as of 2022. These statistics show the importance of research to prevent it. 

CRC starts as a polyp in the inner lining of the colon. Hence, early detection of polyps is the best way to prevent CRC. It is commonly seen in the older population. Most polyps are noncancerous, but some of them can lead to CRC. To avoid it, doctors have to check for the polyps manually. An endoscope with a camera positioned at its tip is inserted in the colon to inspect it. It is called colonoscopy. This is a labor-intensive and time-consuming process. The presence of polyps of different sizes at different locations poses challenges to the doctors during colonoscopy. Distinguishing polyp from the colon body is more difficult with the presence of wrinkles on the colon. These challenges can sometimes lead to failure in the early detection of CRC. It is also labor-intensive and costly. With the advancement of machine learning, researchers are now trying to automate the polpy detection process. The main idea is to detect the polyp directly from the images coming from the camera feed using various deep-learning models. 

Various deep learning models have been employed so far to diagnose CRC and other cancers\cite{pacal2022efficient,bora2021computational,theodosi2021design}. CNN models such as Faster RCNN \cite{qian2020new}, MRCNN
\cite{yang2020colon}, combined U-Nets \cite{thanh2020polyp} and YOLO \cite{ghose2024improved,guo2020reduce,cao2021gastric} based models have shown effectiveness in polyp detection. In this paper, we are trying to see the effectiveness of the recently released YOLOv11 model for polyp detection.

\section{Relevent Works}
YOLO has been one of the most sought-after algorithms recently for object detection, segmentation, and classification tasks. It was first introduced by Redmon et al.\cite{redmon2016you} in 2016. They changed the formulation of the classification problem to a regression problem. The model directly predicts the bounding box coordinates and probability of the class straight from the image pixels. This lowers the complexity of the model with respect to the two-stage models like R-CNN and DPM. The main advantage of YOLO resides in its speed in identifying objects with only one glance. 

Various versions of YOLO have been used for polyp detection. Guo et al.\cite{guo2020reduce} used the YOLOv3 structure with active learning for polyp detection. After retraining the false positive images as negative samples multiple times, they reduced the false positive rate. They achieved an FPR of 1.5\% on colonoscopy videos. Cao et al. \cite{cao2021gastric} could detect small polyp using an integration of feature extraction and fusion module and YOLOv3. Their model fused the semantic information of high-level feature maps with low-level feature maps. Pacal et al. \cite{pacal2022efficient} used augmentation techniques and transfer learning on YOLOv3. They also used the SiLU activation function and complete intersection over Union loss functions for performance improvement. The model was trained and tested on the SUN poly and PICCOLO databases.
The multi-scale mesh based on YOLOv4 was introduced by Lee et al.\cite{lee2022improvement} for effectively detecting small polyps. They used Autoaugment to create images of different scales and achieved an mAP of 98.36. Wan et al.\cite{wan2021polyp} added a self-attention module on the top of the backbone network of YOLOv5. They also used the mosaic method for data augmentation. With this method, they could match the performance of Faster R-CNN in terms of accuracy.  

Qian et al.\cite{qian2022automatic} tried to solve the problem of smaller dataset by using Generative Adversarial Network (CGAN) to create similar images. They also modified YOLOv4 model with dilated convolutions and skip connections to achieve an accuracy of 92.37\% using their expanded dataset. Carrinho et al. \cite{carrinho2023highly} also used YOLOv4 for polyp detection. They extensively analyzed different regularization, data pre-processing, and data augmentation methods. They experimented with FP32, FP16, and INT8 quantization levels. ABC optimization algorithm was used to optimize the hyper-parameters of the YOLO model by Karamn et al.\cite{karaman2023hyper}. They improved more than 2\% in F1-score and 3\% in mAP. Ghose et al. \cite{ghose2024improved} used a finetuned YOLOv5 model. They used data augmentation techniques to increase the number of images and, hence, increase the performance of the model. Mehrshad and ali \cite{lalinia2024colorectal} used a collective dataset consisting of 5 different publicly available datasets, namely: Kvasir-SEG \cite{KVASIR_original_dataset}, CVC-ClinicDB\cite{bernal2015wm}, CVC-ColonDB \cite{tajbakhsh2015automated}, ETIS \cite{silva2014toward}, and EndoScene \cite{vazquez2017benchmark}. They used YOLOv8 for polyp detection and achieved an impressive 95.6\% precision, 91.7\% recall, and 92.4\% F1-score.
 
In this paper, we will use YOLOv11 for polyp detection, which is built upon the YOLOv8 model and showed a higher mAP on the coco dataset using 22\% less parameters. We will use both original and expanded dataset with augmentation to study the performance of 5 different YOLO models. 

\section{Materials and methods}
\subsection{Dataset description}
We have used the Kvasir dataset \cite{KVASIR_original_dataset} for our work. It originally had 1000 images. Images were collected using endoscopic equipment at Vestre Viken Health Trust (VV) in Norway. This dataset consists of annotated images of different sizes 720x576 up to 1920x1072. 

\subsection{Data pre-processing}
In this work, we have used two versions of the Kvasir dataset available in roboflow\cite{kvasir-seg-gouny_dataset}. Initially, we experimented with a simple dataset without augmentation (V6). It had the original 1000 images. It was only resized to 640*640. It was segregated into training (800 images), testing(100 images), and validation (100 images) data. 
We used another version of the KVASIR seg dataset from roboflow (V5) to further increase the model's performance. Augmentation was used in this dataset. Image flipping(horizontal and vertical), 90-degree rotation ( clockwise and counterclockwise, upside down), cropping( zoom between 0\% to 20\%), rotation (between -15\% to +15\%), shear($\pm15\degree$ horizontal and vertical), change of brightness(-25\% to +25\%) and blurring effects have been used. The total number of images after augmentation was 2600. It was further segregated into training (2400 images), testing (100 images), and validation (100 images) datasets.


\subsection{Architecutre of YOLOv11}
YOLOv11 has been released recently with some key architectural improvements. It builds upon the strengths of previous versions and introduces several architectural enhancements. It can be used for pose estimation and instance segmentation in addition to conventional object detection tasks. YOLO models consist of three components: Backbone, neck, and head. YOLOv11 also follows that. It builds upon the YOLOv8 model \cite{khanam2024YOLOv11}. 

\subsubsection{Backbone}
The backbone extracts features from the input image using convolutional layers, Spatial Pyramid Pooling - Fast (SPPF) block, and Cross Stage Partial with Spatial Attention (C2PSA) block. The convolutional layers use Sigmoid-Weighted Linear Units (SiLU) as the activation function. The SPPF block segregates the image into grids and extracts features from them independently to handle multi-scale information using max pooling. It further combines information from different scales. This module also helps in maintaining speed. 
\subsubsection{Neck}

C2PSA block uses two Position-sensitive attention modules. The PSA module processes input tensors along with position-sensitive attention. It helps in focusing selectively on finer details. This module also concatenates the outputs for the input tensor and attention layer. Two PSA modules operating on the separate parts of the feature map are concatenated. 

YOLOv11 uses C3K2 module in place of C2F module of YOLOv8. The C3K2 module has two convolutional layers at the start and end, with multiple C3K modules in between. It employs smaller 3*3 kernels for capturing essential features. It is an improvement of the CSP(Corss Stage Partial) bottleneck from earlier versions. The series use of small kernels to process separate feature maps and merging them after convolution improves the feature representation.    

\subsubsection{Head}

It consists of C3k2 blocks and final convolution layers. It uses multi-scale prediction from three feature maps of three scales—small (P3), medium (P4), and large (P5). It is helpful to detect objects of varying sizes. Ultimately, the final detection layer gives the class predictions and bounding boxes.  

\begin{figure*}[htbp]
\centerline{\includegraphics[width=\textwidth]{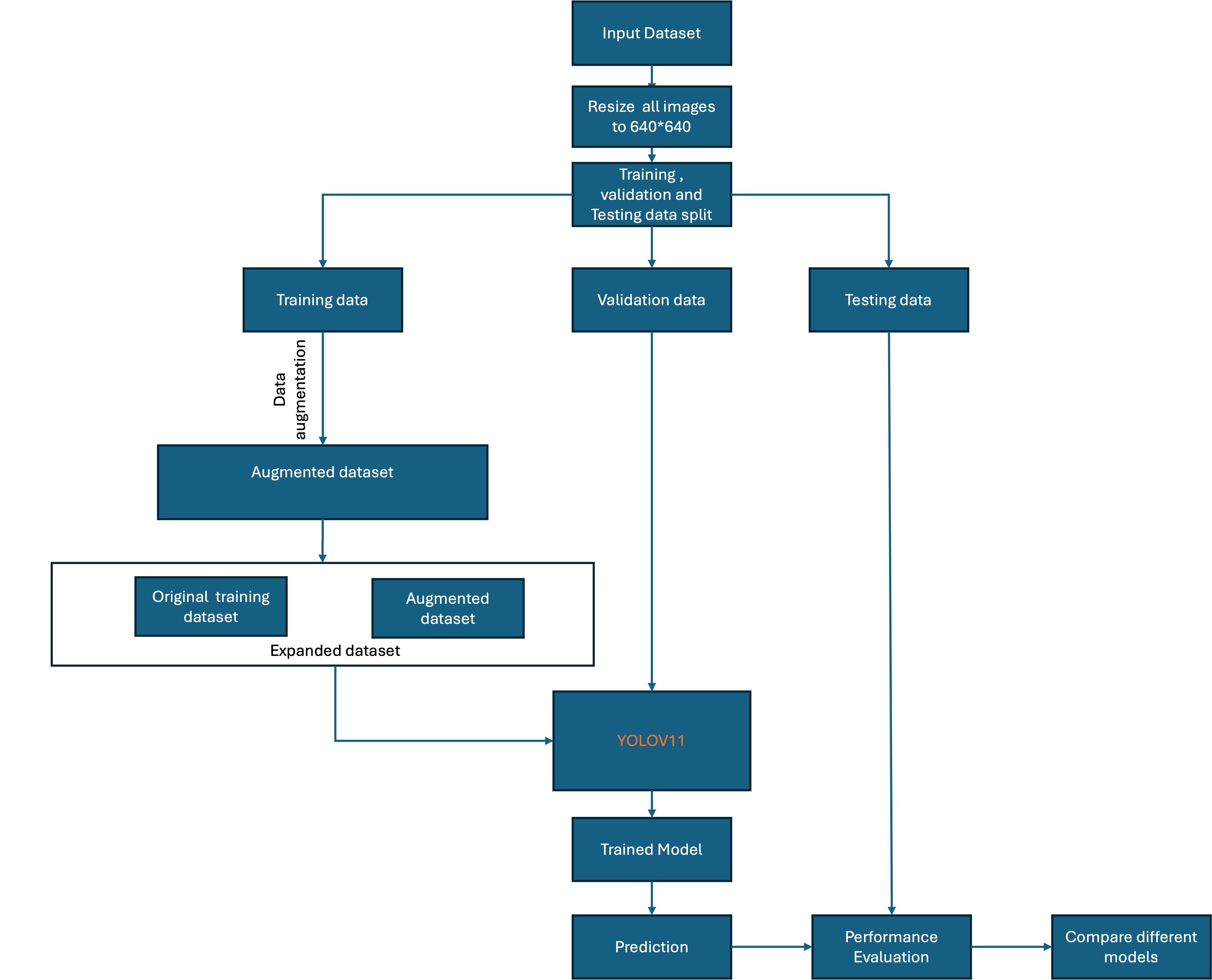}}
\caption{Proposed method for Polyp detection using YOLOv11}
\label{fig1}
\end{figure*}

The main advantages of YOLOv11 are its increased precision with lower complexity, optimized speed, and adaptability in different deployable scenarios such as cloud platforms and edge devices. Figure \ref{fig1} describes the proposed method for polyp detection using YOLOv11

\section{Experiments}
\subsection{Matrices used}
We have used three different matrices to evaluate the performance of the models.
\begin{equation}
 \text{Precision} =\frac{\text{True Positive}}{\text{True Positive + False Positive}}
 \end{equation}

 \begin{equation}
 \text{Recall} =\frac{\text{True Positive}}{\text{True Positive + False Negative}}
  \end{equation}

  \begin{equation}
      \text{F1-score} =\frac{2*\text{Precision}*\text{Recall}}{\text{Precision + Recall}}
   \end{equation}

Here, precision gives the ratio of the actual polyp detected to the detected polyp instances. Higher precision can reduce false polyp detections. Recall, on the other hand, is the ratio of the detected polyp to all the instances. This will show us how effective the model is in detecting polyps among all the instances it was shown. For medical cases, a comparatively higher recall value is desired as it will help in the timely detection of the disease. F1-score sets a balance between these two. It uses both precision and recall to give a score to the model's overall performance. 
\subsection{Experimental setup}
We have used google colab pro+ platform for all the experiments. Runtype of T4 GPU was used. It uses Tesla T4 GPU with 16 GB RAM. It had an Intel(R) Xeon(R) CPU @ 2.20GHz processor, with disk space of 236 GB and RAM of 52 GB.    
\subsection{Training and testing}

YOLO-v11 has five different types of models, namely: YOLO11n, YOLO11s, YOLO11m, YOLO11l, and YOLO11x. All five models were trained on the two dataset versions detailed in the previous section. Training was done for 200 epochs. Early stopping was used with a patience of 20 epochs. AdamW was used as the optimizer. The training parameters used for the training have been detailed in the table \ref{trainingparameters}. For YOLO11x, a batch size of 8 was used to avoid the constraint of GPU memory in the Google colab.


\begin{table}[htbp]
\caption{Parameters Used for training }
\begin{center}
\begin{tabular}{|c|c|}
\hline
\textbf{No. of epochs} & 200 \\
\hline
\textbf{Optimizer } & AdamW  \\
\hline
\textbf{Learning rate } & 0.002 \\
\hline
\textbf{Batch size } & 16 \\
\hline

\end{tabular}
\label{trainingparameters}
\end{center}
\end{table}

\section{Results and discussion}

\begin{figure}
    \includegraphics[width=.12\textwidth]{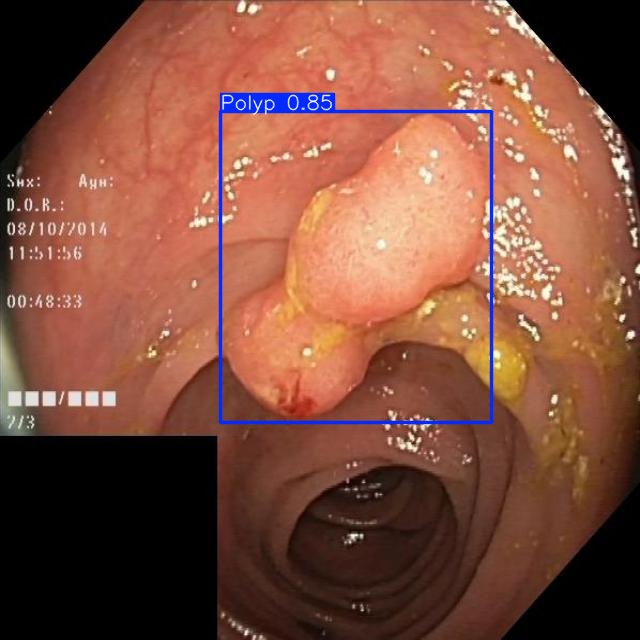}\hfill
    \includegraphics[width=.12\textwidth]{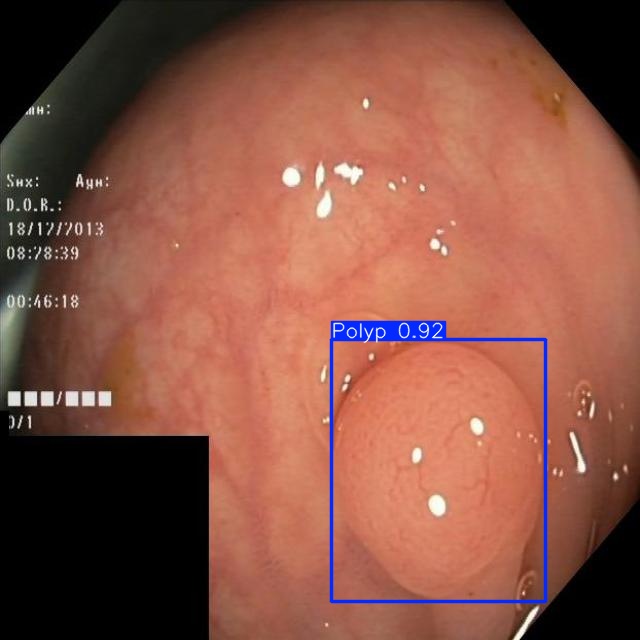}\hfill
    \includegraphics[width=.12\textwidth]{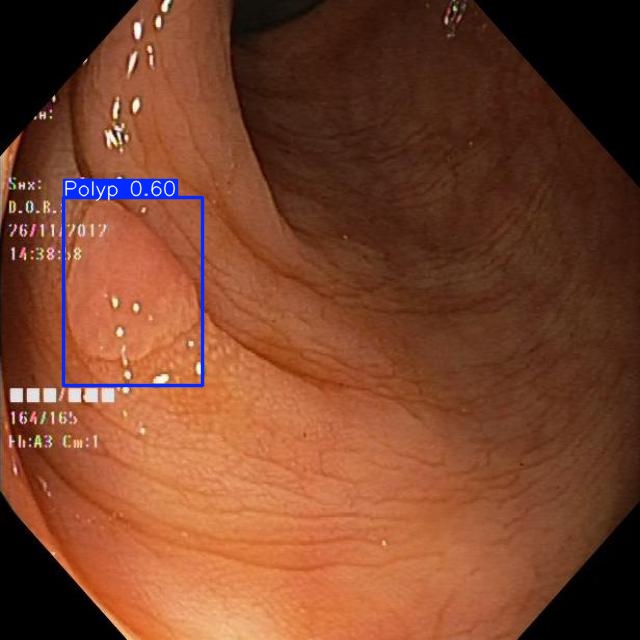}\hfill
    \includegraphics[width=.12\textwidth]{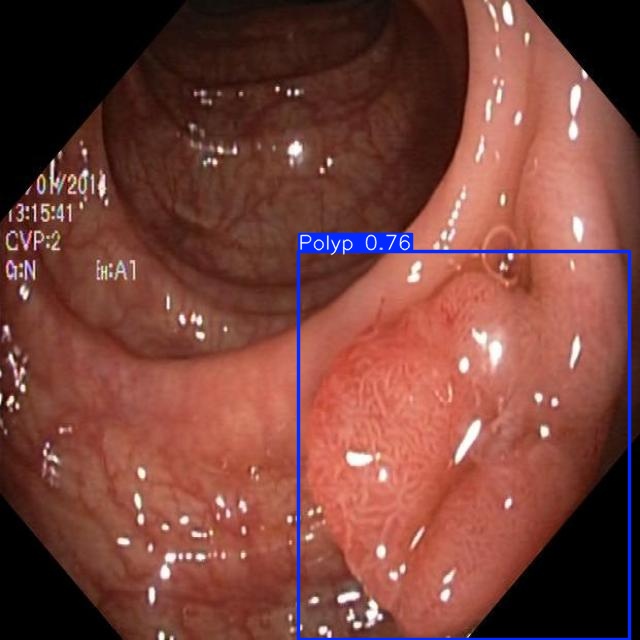}
    \\[\smallskipamount]
    \includegraphics[width=.12\textwidth]{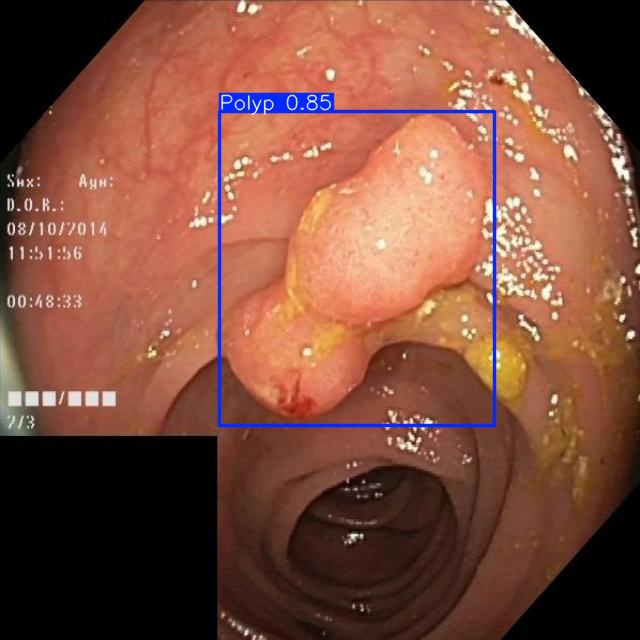}\hfill
    \includegraphics[width=.12\textwidth]{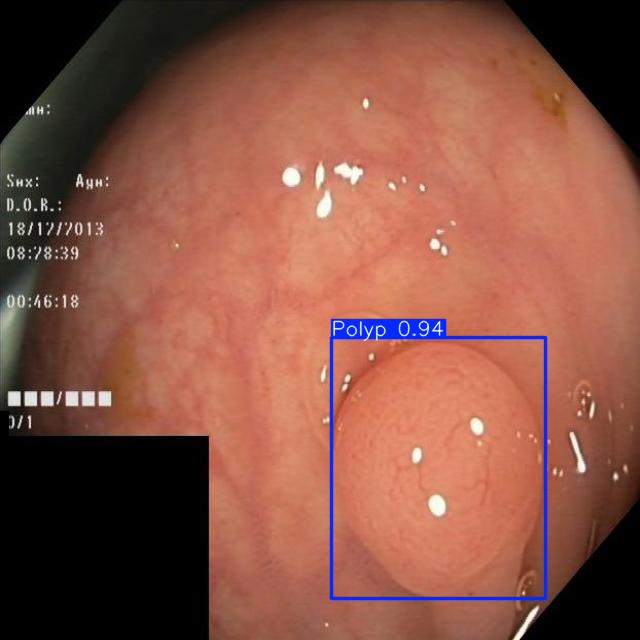}\hfill
    \includegraphics[width=.12\textwidth]{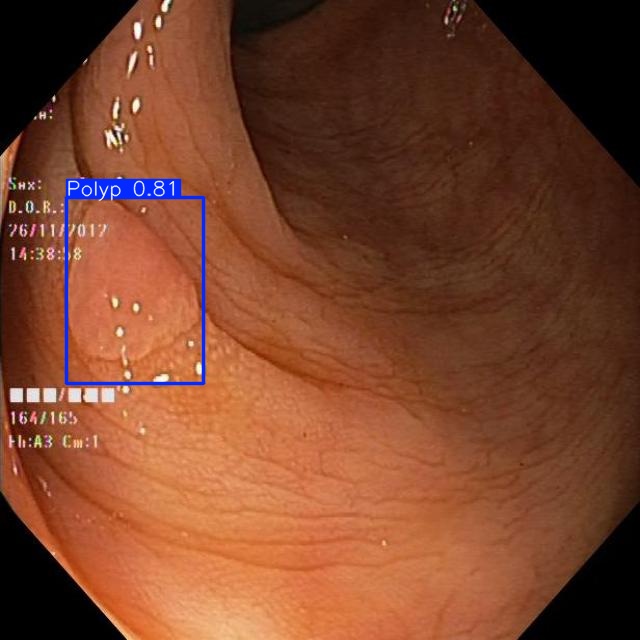}\hfill
    \includegraphics[width=.12\textwidth]{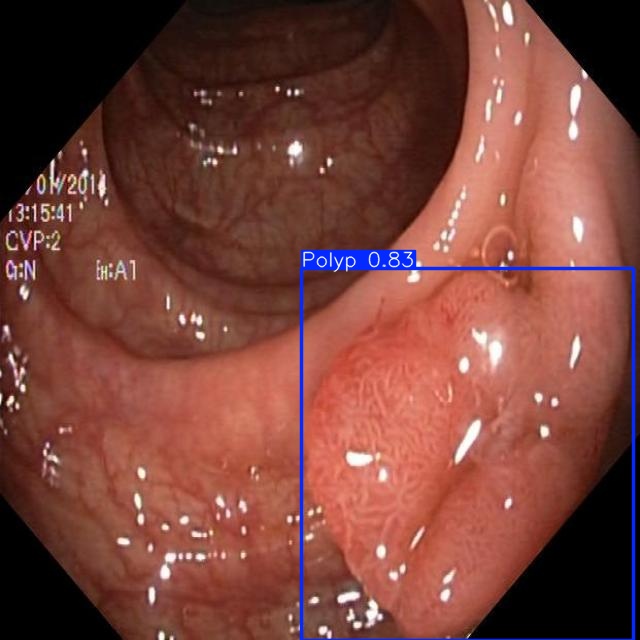}
    \\[\smallskipamount]
    \includegraphics[width=.12\textwidth]{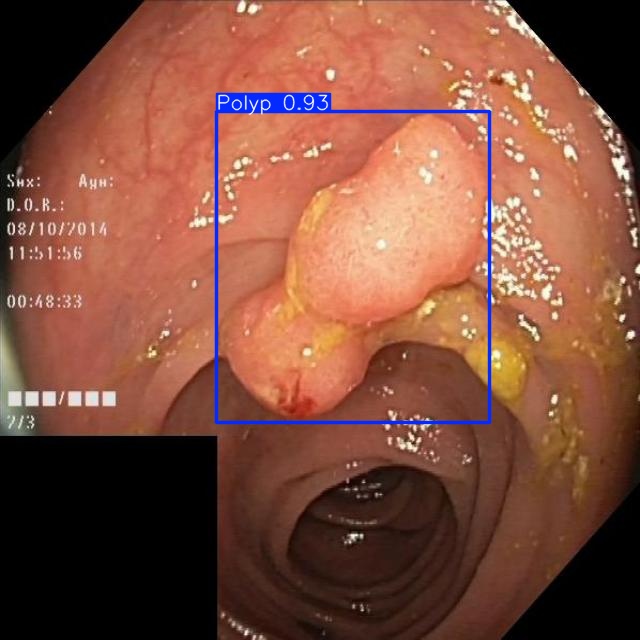}\hfill
    \includegraphics[width=.12\textwidth]{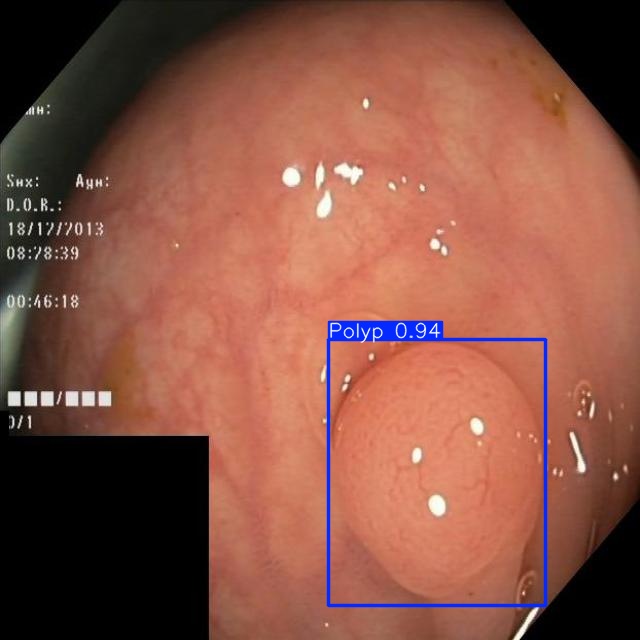}\hfill
    \includegraphics[width=.12\textwidth]{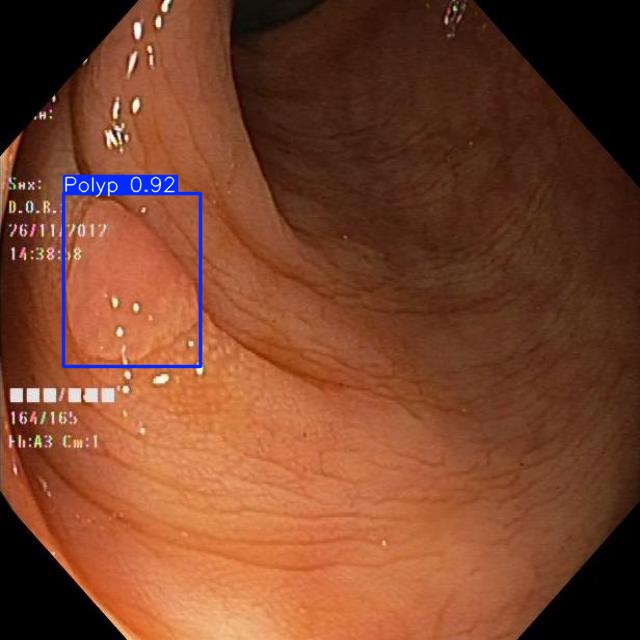}\hfill
    \includegraphics[width=.12\textwidth]{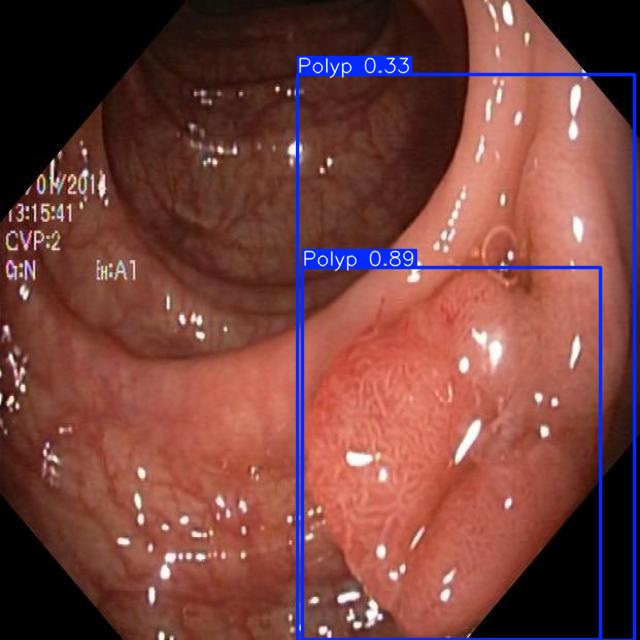}
    \\[\smallskipamount]
    \includegraphics[width=.12\textwidth]{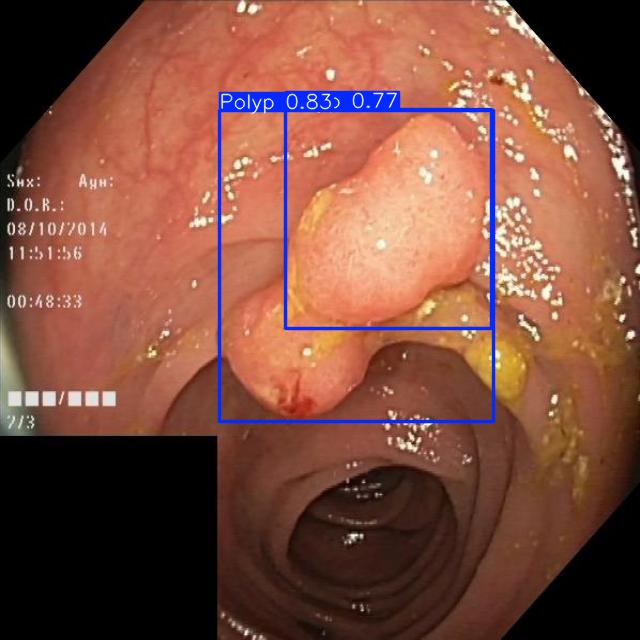}\hfill
    \includegraphics[width=.12\textwidth]{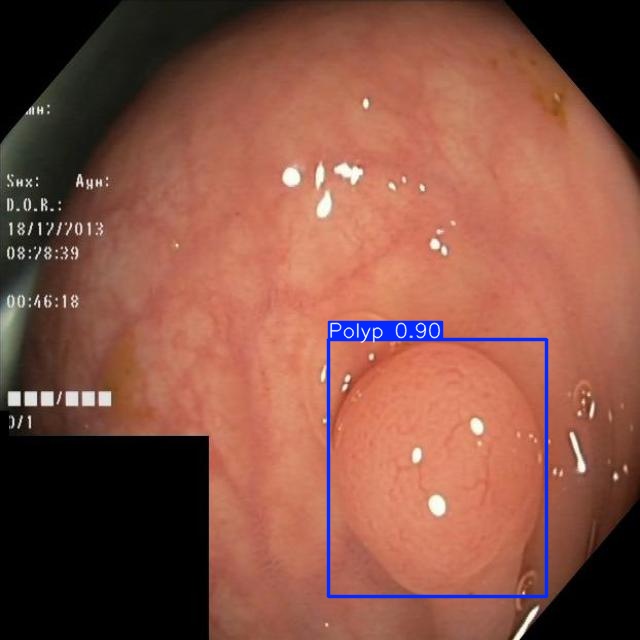}\hfill
    \includegraphics[width=.12\textwidth]{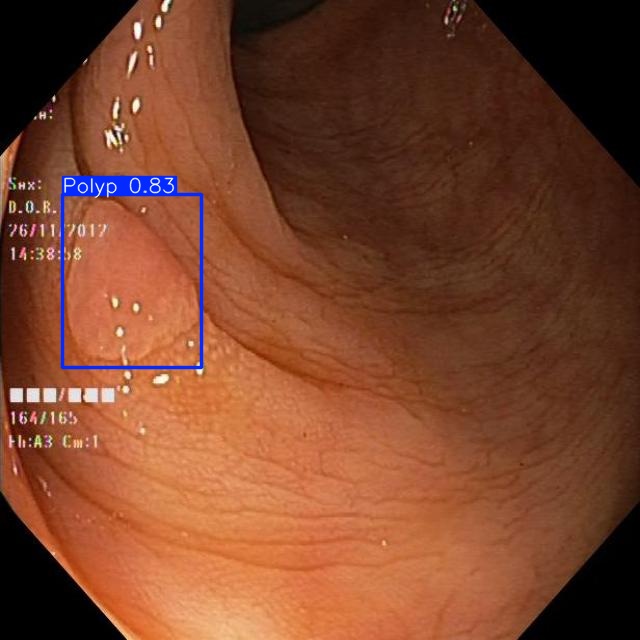}\hfill
    \includegraphics[width=.12\textwidth]{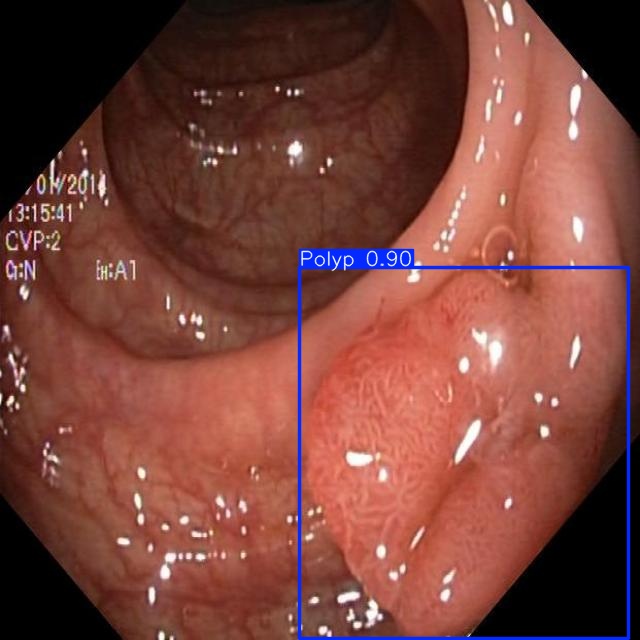}
    \\[\smallskipamount]
    \includegraphics[width=.12\textwidth]{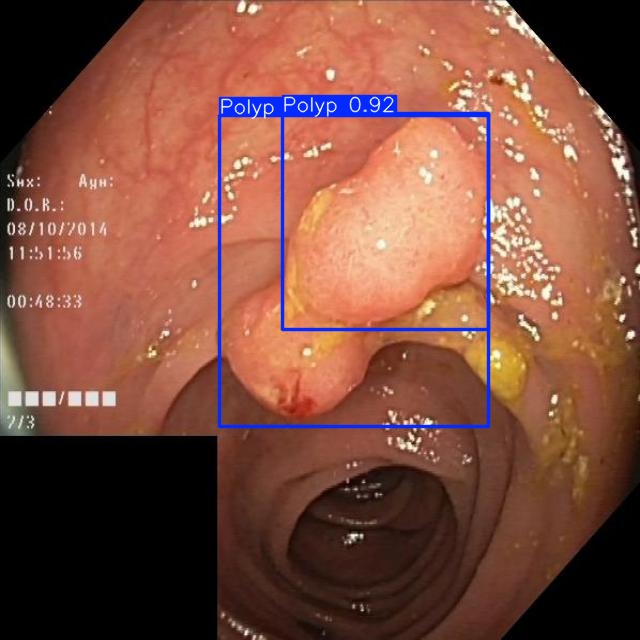}\hfill
    \includegraphics[width=.12\textwidth]{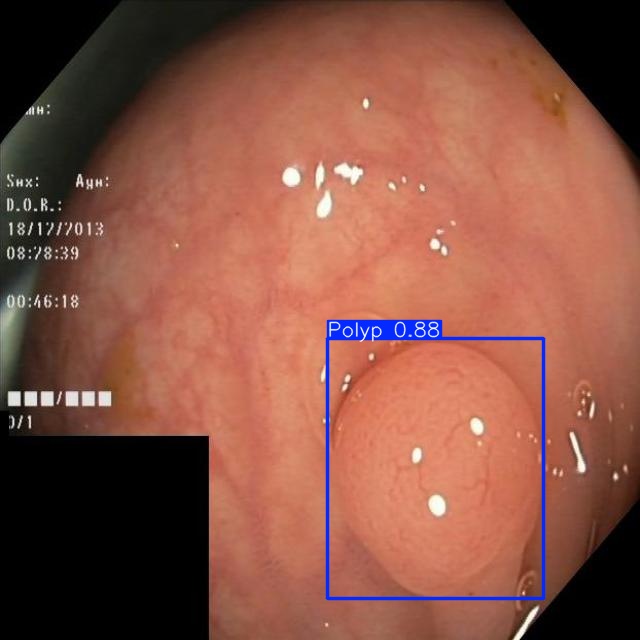}\hfill
    \includegraphics[width=.12\textwidth]{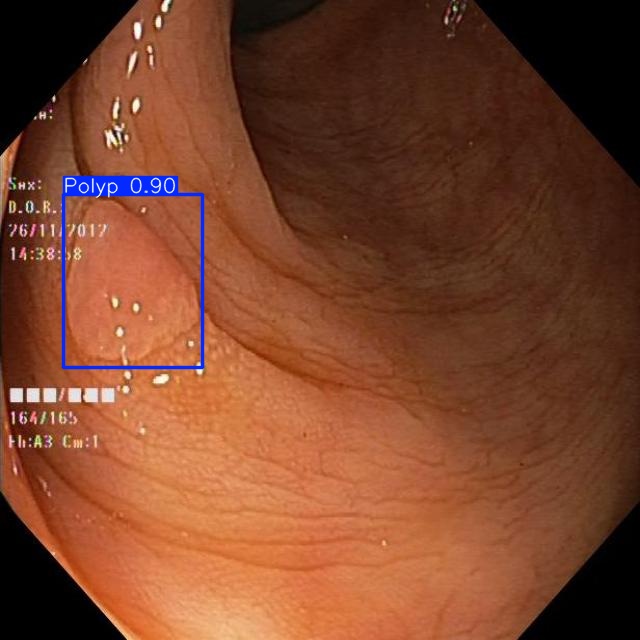}\hfill
    \includegraphics[width=.12\textwidth]{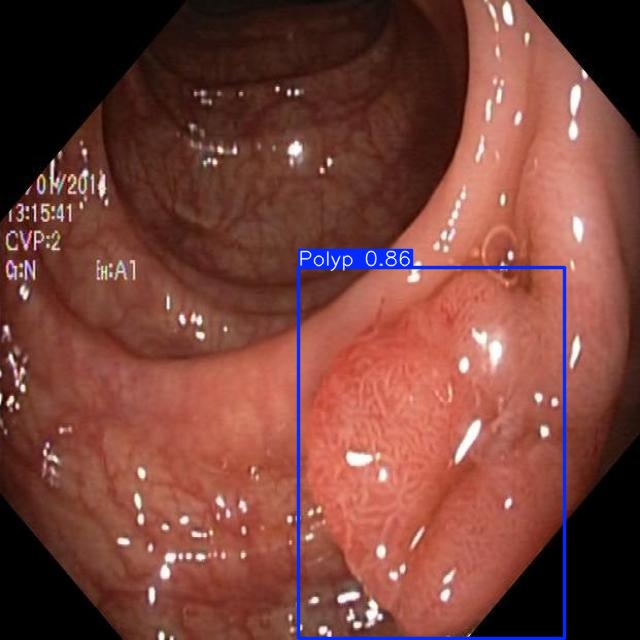}
    \\[\smallskipamount]
    \caption{Polyp detection with (a) YOLO11n (b) YOLO11s (c) YOLO11m (d) YOLO11l (e) YOLO11x with the model trained with original images}\label{Yolowithoriginalimage}
\end{figure}

\begin{figure}[htbp]
\centerline{\includegraphics[width=0.5\textwidth]{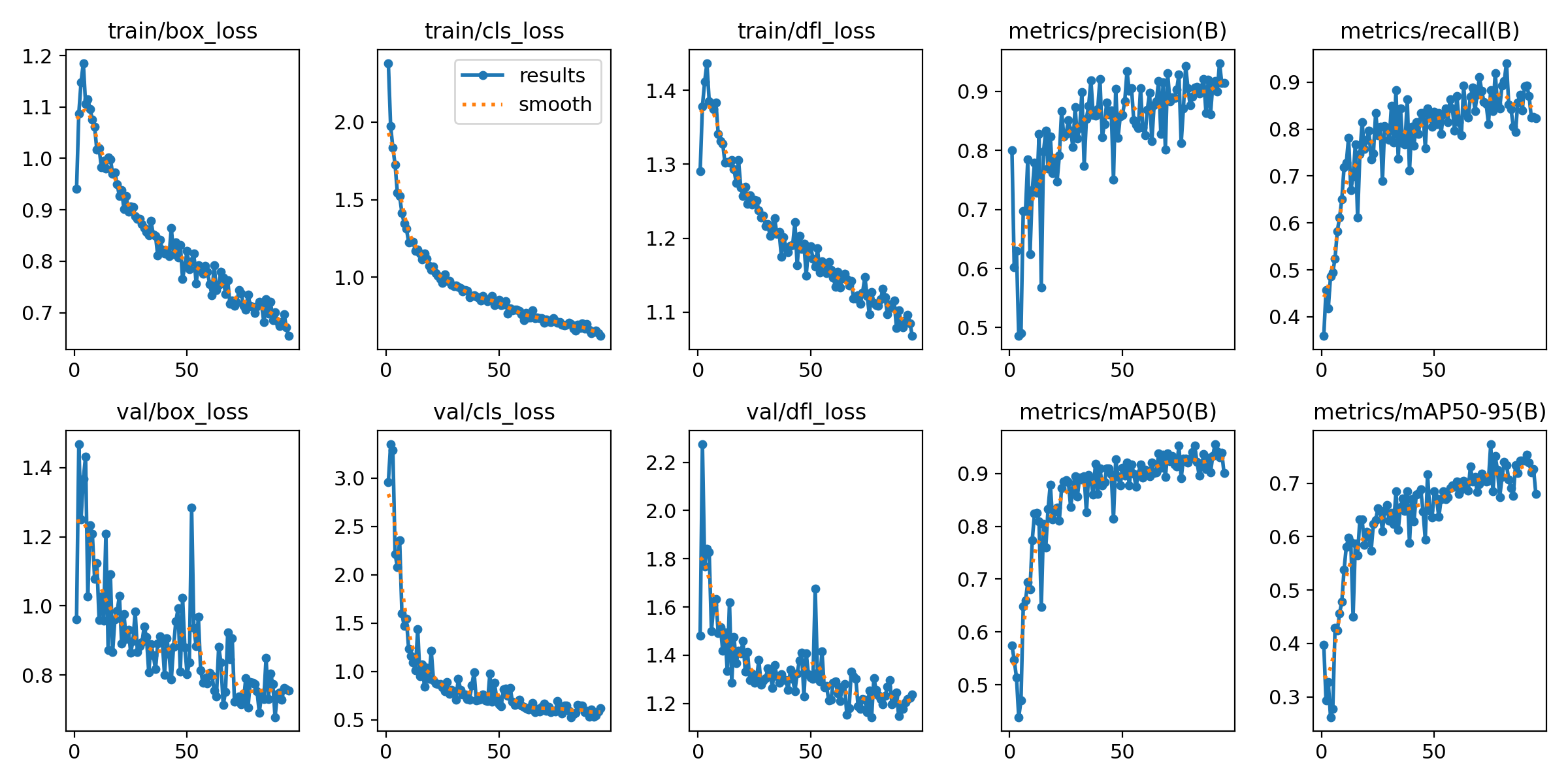}}
\caption{ Different losses on training and validation
dataset during model training YOLOv11n model with original dataset }
\label{losses}
\end{figure}

We first experimented with the original dataset consisting of 1000 images. The results obtained for each model has been described in the table \ref{YOLOv11_1kdata}. Fig \ref{Yolowithoriginalimage} shows the prediction result of the YOLO models trained with the original dataset. Fig\ref{losses} shows different losses in training and validation for YOLO11n.
The table shows that YOLO11l has the highest F1 score, which is slightly (0.95\%) more than YOLO11n. The recall of YOLO11l 4.85\% more, but its precision is 2.77 \% lower. A slightly higher (0.95\%) F1 score was achieved by YOLO11l using approximately 9 times more parameters than YOLO11n. The F1-score for YOLO11x was the lowest among all five models. The inferior performance of the model can be attributed to the smaller dataset available for training such a big model with 56.9M parameters. This motivated us to improve the model's performance by training the models with larger dataset. It can be done using augmentation techniques such as image flipping, rotation, blurring, etc. 

\begin{table}[htbp]
\caption{Results for the training of 5 different versions of YOLO-V11 }
\begin{center}
\begin{tabular}{|p{41pt}|p{27pt}|p{20pt}|p{25pt}|p{35pt}|p{32pt}|}
\hline
\textbf{YOLOv11 Model} & \textbf{\textit{Precision}}& \textbf{\textit{Recall}}& \textbf{\textit{F1-score}}& \textbf{\textit{Parameters (M)}}& \textbf{\textit{Training time}} \\
\hline
\textbf{YOLO11n } & \textbf{0.9176} & 0.9029 & 0.9101 &2.6&29m 27s\\
\hline
\textbf{YOLO11s } & 0.8558 & 0.9223 & 0.8878 & 9.4 & 29m 43s \\
\hline
\textbf{YOLO11m } & 0.8521 & \textbf{0.9514} & 0.8990 & 20.1 &110m 12s \\
\hline
\textbf{YOLO11l } & 0.8899 & \textbf{0.9514} & \textbf{0.9196} & 25.3 & 92m 11s\\
\hline
\textbf{YOLO11x } & 0.8245 & 0.9126 & 0.8663 & 56.9 &143m 44s\\
\hline
\end{tabular}
\label{YOLOv11_1kdata}
\end{center}
\end{table}

We used KVASIR seg dataset from roboflow (V5) for that. The training was done for 200 epochs with a patience of 20, similar to the previous one. Table \ref{YOLOv11resultswithaugmentation} details the results obtained from all five models of YOLOv11 using the expanded dataset. YOLO11n has the highest precision (0.9056) and F1 score (0.9326), but YOLO11m has the highest Recall score (0.9514). The recall and F1 scores of YOLO11n have increased by 3.22\% and 0.93\%, respectively, with respect to the model trained with the original images. YOLO11m model has the highest recall score of 0.9514, but its precision (0.8596) is inferior than YOLO11n. The YOLO11n model performed better with expanded dataset than the original dataset. Fig \ref{YOLOv11resultswithaugmentation} shows the prediction results using all five models. Considering the results from both versions of the dataset and the number of parameters used, we can say that YOLO11n is performing the best for polyp detection.

\begin{table}[htbp]
\caption{Results for the training of 5 different versions of YOLO-V11 with augmented dataset}
\begin{center}
\begin{tabular}
{|p{41pt}|p{27pt}|p{20pt}|p{25pt}|p{35pt}|p{32pt}|}

\hline
\textbf{YOLOv11 Model} & \textbf{\textit{Precision}}& \textbf{\textit{Recall}}& \textbf{\textit{F1-score}}& \textbf{\textit{Parameters (M)}}& \textbf{\textit{Training time}} \\

\hline
\textbf{YOLO11n } & \textbf{0.9056} & 0.9320 & \textbf{0.9186} & 2.6 & 157m 22s\\
\hline
\textbf{YOLO11s } & 0.8434 & 0.9417 & 0.8898 & 9.4 & 108m 54s \\
\hline
\textbf{YOLO11m } & 0.8596 & \textbf{0.9514} & \textbf{0.9186} & 20.1 & 165m 6s \\
\hline
\textbf{YOLO11l } & 0.8508 & 0.9417 & 0.8939 & 25.3 & 180m 12s \\
\hline

\textbf{YOLO11x } & 0.6713 & 0.9320 & 0.7804 & 56.9 & 121m 27s \\
\hline
\end{tabular}
\label{YOLOv11resultswithaugmentation}
\end{center}
\end{table}

\begin{figure}

    \includegraphics[width=.12\textwidth]{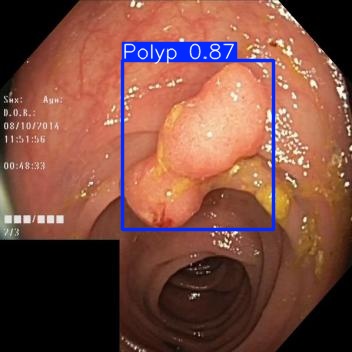}\hfill
    \includegraphics[width=.12\textwidth]{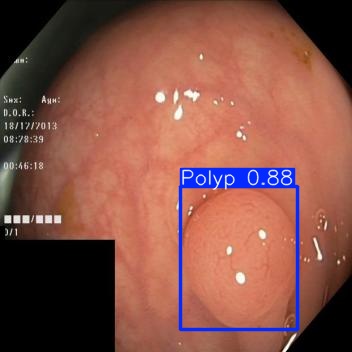}\hfill
    \includegraphics[width=.12\textwidth]{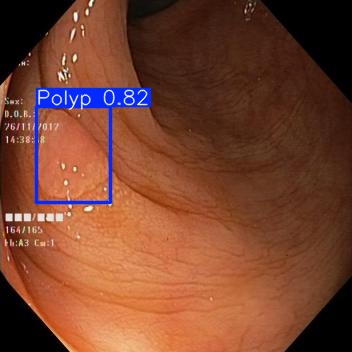}\hfill
    \includegraphics[width=.12\textwidth]{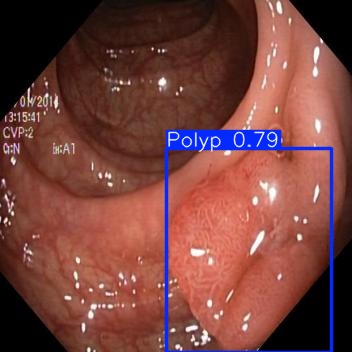}
    \\[\smallskipamount]
    \includegraphics[width=.12\textwidth]{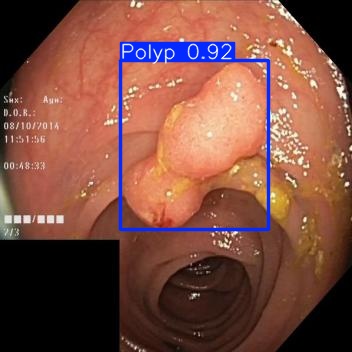}\hfill
    \includegraphics[width=.12\textwidth]{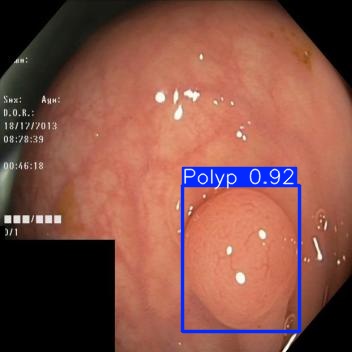}\hfill
    \includegraphics[width=.12\textwidth]{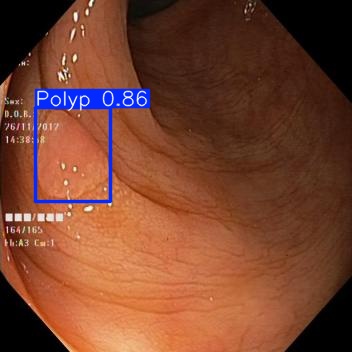}\hfill
    \includegraphics[width=.12\textwidth]{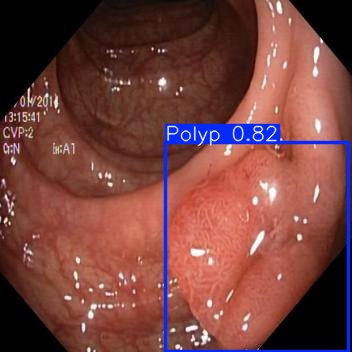}
    \\[\smallskipamount]
    \includegraphics[width=.12\textwidth]{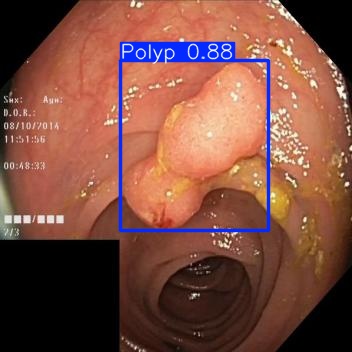}\hfill
    \includegraphics[width=.12\textwidth]{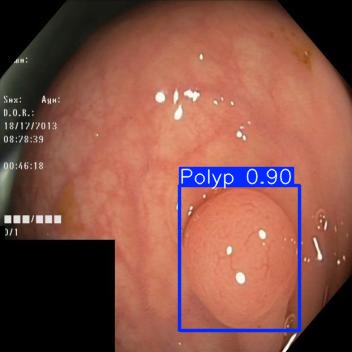}\hfill
    \includegraphics[width=.12\textwidth]{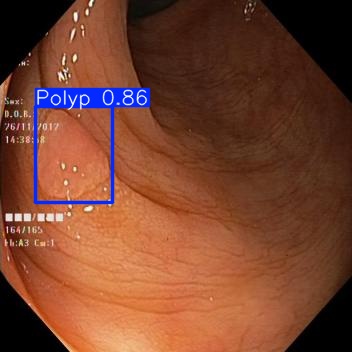}\hfill
    \includegraphics[width=.12\textwidth]{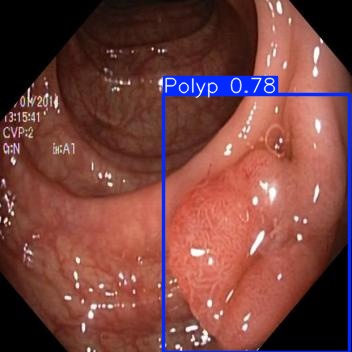}
    \\[\smallskipamount]
    \includegraphics[width=.12\textwidth]{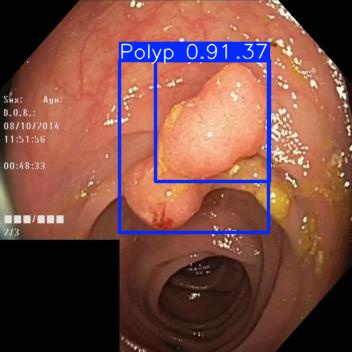}\hfill
    \includegraphics[width=.12\textwidth]{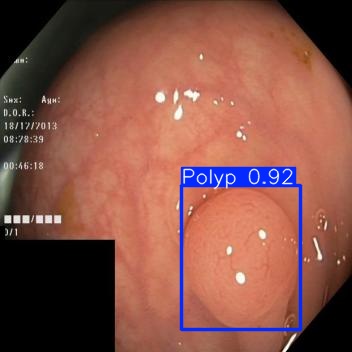}\hfill
    \includegraphics[width=.12\textwidth]{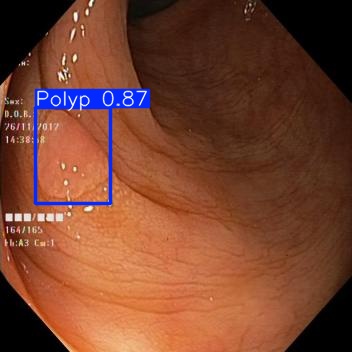}\hfill
    \includegraphics[width=.12\textwidth]{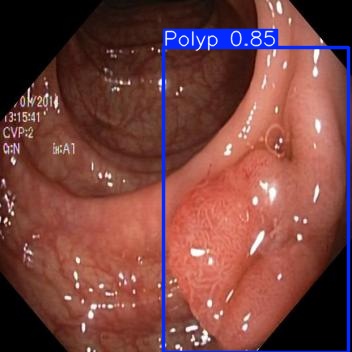}
    \\[\smallskipamount]
    \includegraphics[width=.12\textwidth]{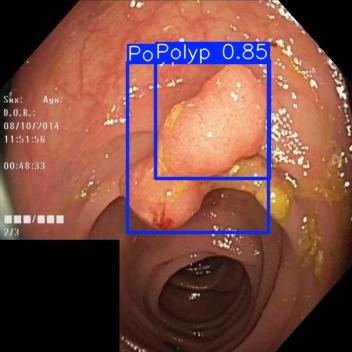}\hfill
    \includegraphics[width=.12\textwidth]{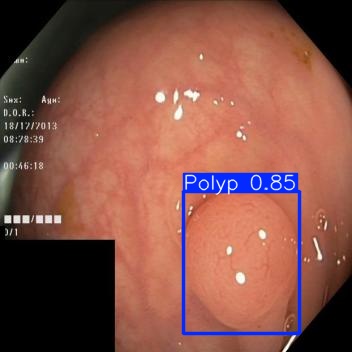}\hfill
    \includegraphics[width=.12\textwidth]{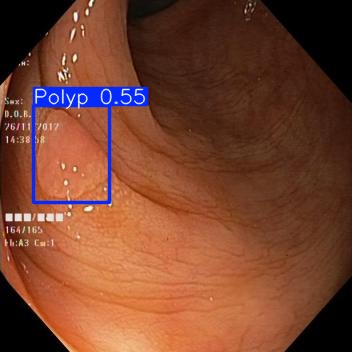}\hfill
    \includegraphics[width=.12\textwidth]{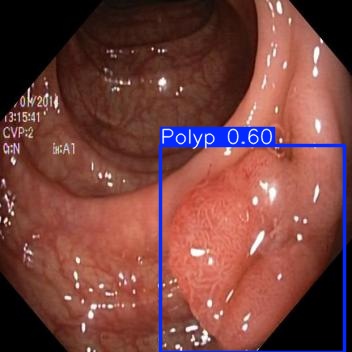}
    \\[\smallskipamount]
    \caption{Polyp detection with (a) YOLO11n (b) YOLO11s (c) YOLO11m (d) YOLO11l (e) YOLO11x  with the model trained with the augmented dataset}\label{fig:foobar}
\end{figure}

\section{conclusion and future scope}
In this paper, we tried to analyze the effect of YOLOv11's five models for the polyp detection task. We used two different versions of KVASIR seg dataset from roboflow. The first had 1000 original images, and the second had 2400 images. Augmentation techniques were used to enhance the number of images in the second one. We tested both versions of the dataset to train and test 5 different versions of YOLOv11 (YOLO11n, YOLO11s, YOLO11m, YOLO11l, YOLO11x). YOLO11n performed reasonably well in both datasets if we consider the F1 score with respect to the number of parameters used. In the augmented dataset, the recall and F1-score improved by 3.2\% and 0.93\%, respectively, but precision was reduced by 1.3\%. As improving recall while maintaining reasonable precision is vital in medical imaging, we can safely say YOLO11n can be used for polyp detection. It's lightweight and superior performance makes it an ideal model among all five to be used in cloud and edge devices. As evident from the experiment, the model's performance can be improved by augmentation. In the future, we will explore various optimization techniques to improve the model's performance.








\bibliographystyle{ieeetr}
\bibliography{references.bib}

\end{document}